# Human Centric Embodied Intelligence for Soft Wearable Robotics

*Rainier Natividad*[1], *Raye Chen-Hua Yeow*[1]

[1] Department of Biomedical Engineering, National University of Singapore

### *Abstract*

Soft wearable robots have evolved rapidly from proof-of-concept devices into promising platforms for rehabilitation, occupational assistance, and human augmentation. As the field matures, its central challenge extends beyond the development of softer materials and more capable actuators to the integration of sensing, intelligence, and human adaptation into systems that users can wear comfortably, trust, and benefit from over extended periods. This transition motivates the concept of Human-Centric Embodied Intelligence (HCEI), in which intelligence emerges from the coupled human–robot system through the interaction of morphology, multimodal sensing, adaptive cognition, compliant actuation, and the wearer's own physiological and behavioral adaptation. To organize this perspective, this review introduces the Perception-Cognition-Actuation-Augmentation (PCAA) framework, which positions perception and cognition as the primary drivers of design, shifting development beyond the conventional actuator-first paradigm. Using this framework, the review synthesizes advances in soft materials, wearable sensing, artificial intelligence, actuation, human–robot interaction, digital twins, clinical translation, manufacturing, regulation, and ethics, highlighting how these interdependent components collectively shape long-term personalization and real-world deployment. By providing a unified conceptual framework and design perspective, this review aims to guide future research, foster interdisciplinary collaboration, and accelerate the translation of next-generation soft wearable robots toward personalized, predictive, and human-centric wearable intelligence.

# 1. Introduction and Evolution of Soft Wearable Robotics

Wearable robotics has matured into a principal response to human needs that fixed laboratory and clinical equipment cannot meet, because its defining promise is to deliver robotic assistance directly on the body and during natural movement, beyond the bounds of a fixed workspace [1]. The motivating demand is broad and demographically entrenched, and in rehabilitation it is anchored by the burden of neurological injury, of which stroke alone accounts for tens of millions of incident cases and a large share of acquired adult disability each year, leaving survivors with the gait and upper-limb deficits worn assistance is designed to address [2]. Occupational settings are an equally consequential driver, where passive and active wearables reduce the lumbar loading of repetitive lifting and the shoulder loading of sustained overhead work, targeting the cumulative musculoskeletal injuries that dominate industrial health records [3], [4]. A third driver is augmentation of the unimpaired body, where portable exosuits lower the metabolic cost of walking and running [5]. Across these applications a single thread recurs, in that the field is shifting from a fixed assistive force toward platforms that sense, interpret, and co-adapt with an individual user over time, made feasible by advances in soft materials, compliant actuation, flexible on-body sensing, and conformal electronics [6].

Soft wearable robots are the natural substrate for this shift. Rigid exoskeletons remain the established archetype of worn assistance and deliver substantial, well-controlled joint support, but an external rigid frame brackets the limb, introducing bulk, joint misalignment between mechanism and biological articulation, and a discomfort that compounds over daily wear [7]. Soft systems invert these priorities, anchoring forces through compliant materials, textiles, and body-conformal interfaces that load the body in parallel with muscle and tendon; their value extends beyond material softness and rests on a distinct philosophy of human–robot physical integration in which the device conforms to the body as an integrated layer [1]. This review takes that shift as its organizing theme and argues that the field's next stage is best understood through the paradigm of human-centric embodied intelligence (HCEI). In this paradigm, the wearer's body, the soft device, and their continuous co-adaptation form a single coupled system whose intelligence is distributed across materials, morphology, sensing, learning, and the wearer's own adaptation. Figure 1 illustrates this concept as the field's movement from device-centric mechatronic assistance toward HCEI.

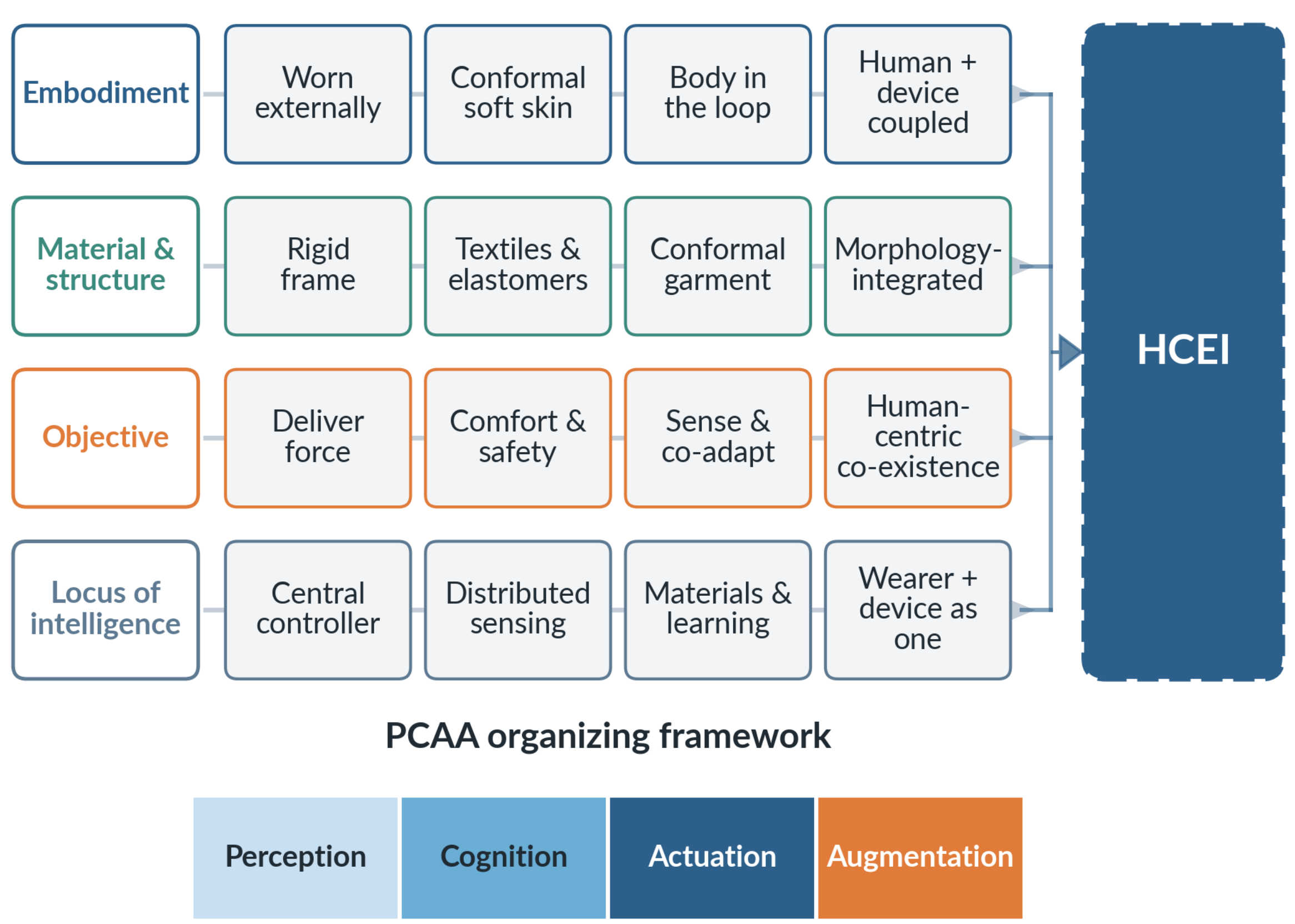

**Fig. 1. Evolution toward Human-Centric Embodied Intelligence.** Four converging dimensions of change (embodiment, material architecture, optimization objective, and locus of intelligence) are shown as parallel tracks progressing from device-centric mechatronic assistance (left) toward the proposed HCEI paradigm (right, dashed). The PCAA framework (Perception–Cognition–Actuation–Augmentation) is shown as the cross-cutting analytical structure through which all four tracks are read. The HCEI endpoint is a forward-looking synthesis rather than a realized class.

### *1.1 Scope and Contributions*

This integrative, narrative review does not claim systematic-review status; it selects and synthesizes representative primary and review literature to build a conceptual argument, and does not claim protocol-driven or exhaustive coverage. Its inclusion logic follows the field's functional span (textile exosuits, soft exoskeletons, robotic gloves, and soft–rigid hybrids, together with their on-body sensing and artificial intelligence and their translation into rehabilitation, occupational assistance, and human augmentation), while purely rigid exoskeletons, prosthetics, and implantable robots enter only where they supply historical context or conceptual contrast [7]. Numerous high-quality reviews already exist, and the gap addressed here is integration: existing surveys are typically organized around device class, body region, actuation method, or clinical application, and no single account yet connects the field's historical evolution, soft embodiment, on-body artificial intelligence, personalized user representations, digital twins, shared human–robot intelligence, and translational readiness within one continuous narrative [8], [9].

The review makes three contributions. First, it advances HCEIas a paradigm for the field (a lens that treats the wearer, the soft device, and their continuous co-adaptation as one coupled system), and situates that paradigm against the historical evolution from rigid mechatronic assistance out of which the need for such a paradigm arises. Second, it proposes PCAA (Perception–Cognition–Actuation–Augmentation) as the organizing framework for synthesizing advances in soft materials, actuation, sensing, on-body artificial intelligence, human–robot interaction, and clinical translation. The subordinate constructs (a unified four-axis taxonomy, the Human Digital Signature, and the human digital twin as a route to scalable personalization at deployment) serve as section-level tools within this analytical spine. Third, it identifies the coupled grand challenges that gate real-world deployment and proposes a capability roadmap, with forward-looking elements explicitly labelled as projections. Each original construct is labelled as a proposed organizing device at first mention.

HCEI is this review's central thesis: a field-level paradigm, or lens, for what soft wearable robotics should become, not a design procedure. PCAA is the analytical framework built on that lens, and it is the only construct in the paper the word *framework* denotes; the sections that follow are each organized as one of its four pillars. The remaining constructs (the unified taxonomy, the Human Digital Signature, the wearable intelligence stack, the actuator comparison matrix, the clinical readiness scorecard, and the five-generation roadmap) are introduced within a single section to support the HCEI–PCAA spine.

### *1.2 Review Scope and Literature Search*

Consistent with its integrative, narrative character, this review synthesizes a purposively selected body of literature and does not assemble the exhaustive, protocol-driven corpus of a systematic review, and it makes no claim of PRISMA compliance. Sources were identified through IEEE Xplore, PubMed, Scopus, Web of Science, and Google Scholar, supplemented by backward and forward citation tracing from key reviews and seminal primary studies. Representative search terms combined device and domain descriptors (*soft exosuit*, *soft exoskeleton*, *soft wearable robot*, *robotic glove*, *textile actuator*, *exosuit*) with function and application descriptors such as *embodied intelligence*, *wearable sensing*, *intent estimation*, *human-in-the-loop optimization*, *shared control*, *digital twin*, *rehabilitation*, *gait assistance*, and *occupational exoskeleton*. Studies were included when they addressed soft or soft–rigid hybrid wearable systems and their sensing, control, augmentation, or clinical translation. Priority was given to peer-reviewed primary research and authoritative reviews that frame the field's trajectory. Purely rigid exoskeletons, prosthetics, implantable robots, and non-wearable soft robots were excluded except where they supplied historical context or conceptual contrast; non-peer-reviewed sources were limited to a small number of preprints reporting otherwise-unavailable results. The literature spans roughly 1995 to 2026, with foundational works retained for historical grounding and emphasis placed on developments of the past decade. Selection followed conceptual relevance under a purposive process, and the resulting synthesis offers a structured interpretation of the field and does not attempt a quantitative meta-analysis; the attendant limitations are stated explicitly in Section 9.

### *1.3 Four Eras of Wearable Robotics*

Wearable robotics has advanced less through a linear accumulation of devices than through a succession of shifting optimization targets, each generation inheriting its predecessor's hardware and re-optimizing the assistive stack around a newly dominant objective. The progression resolves into four overlapping eras (Fig. 2), a periodization that is an interpretive lens whose boundaries follow the dominant optimization objective and the degree to which the human–robot pairing was treated as one coupled system. Era I (≈1960–2000) engineered rigid, powered exoskeletons for force amplification and the mechanical restoration of lost function, from end-effector neurorehabilitation robots [10] and load-carrying frames [11] to clinical exoskeletons that restored ambulation after motor-complete spinal cord injury [12]; it proved such robots feasible while exposing wearability, embodiment, and acceptance as the unsolved problems organizing what came next. Era II (≈2000–2015) reframed the question from how much torque a worn machine could generate to how compliantly and safely it could be worn, anchoring forces through garment-like textile and elastomer interfaces, as in biologically inspired soft exosuits and the unpowered ankle exoskeleton that reduced metabolic cost with only a spring and clutch; intelligence stayed limited and largely pre-scripted [13], [14]. Era III (≈2015–2025) was the integration phase, re-optimizing soft morphology around perception, control, and clinical validation as soft exosuits and gloves migrated from laboratory demonstration to controlled trials and the sensing-and-control stack thickened toward intent detection and learning [15], [16].
Era IV (≈2025–2040) is prospective, projecting the transformation of soft wearable robots from device-centric products into personalized, sustainable, human-centric embodied systems integrated with high-fidelity user models and on-body artificial intelligence, as the binding constraint shifts toward long-horizon co-existence with a particular body and life. A representative device-by-device timeline is provided as Supplementary Material; the constructs that formalize this direction are developed in the sections that follow.

Historical evolution of wearable assistance

| | Era I | Era II | Era III | *Prospective* Era IV |
|---|---|---|---|---|
| | Mechanical assistance | Soft-robotics revolution | Intelligent wearables | Human-centric embodied intelligence |
| | 1960–2005 | 2005–2015 | 2015–2025 | *2025 →* |
| Dominant objective | Force amplification | Wearability & compliance | Context-aware adaptation | Human-centric co-existence |
| Exemplar system | HAL [133] | Soft exosuit [13] | Ankle exosuit [15] | Sustainable wearables [134] |
| Enabling technologies | Rigid frames; torque control | Textiles; soft pneumatics | Multimodal sensing; learning control | On-body AI; user models |
| Deployment milestone | Feasibility & therapy | Metabolic-cost reduction | Functional field trials | Projected all-day use |

**Fig. 2. Historical evolution of soft wearable robotics across four overlapping eras.** Era boundaries reflect shifts in dominant optimization objective rather than strict chronological breaks; Era IV (~2025 onward) is prospective. The figure provides the field's timeline as context.

## 2. Human-Centric Embodied Intelligence: A New Paradigm

The migration of the field's binding constraint from task assistance toward long-horizon co-existence between a particular body and a particular machine cannot be navigated with a vocabulary borrowed unmodified from general artificial intelligence or general soft robotics, because neither was built around a worn, adaptive human partner. The review therefore requires a precise, wearable-specific construct with enough analytical content to organize the field: human-centric embodied intelligence (HCEI). The argument moves from the embodied-intelligence view inherited from soft robotics, narrows it to the coupled human–robot ensemble a wearable creates, states a working definition, decomposes it into principles that also serve as evaluation criteria, and contrasts the construct with the paradigms it subsumes. HCEI is presented as the synthesizing lens of this review; its extensions beyond the cited literature are flagged as proposals that remain open to validation.

## 2.1 From Intelligence in Software to Body–Brain–Environment Loops

Embodied intelligence locates intelligence in the interaction of physical body, controller, and environment, moving beyond a controller-centered view that treats the body as a mere effector [17]. The morphological-computation thesis sharpens this, in that a compliant body subjected to forces offloads, through its own passive dynamics, part of the information processing a rigid system would delegate to its controller, making material distribution, geometry, and stiffness active parts of the system's intelligence budget [18]. Passive-dynamic walkers furnish the canonical demonstration, achieving stable, efficient locomotion with little active control by encoding the gait in limb morphology and body–ground mechanics [19]. The same logic has been extended to wearable robotics with the unpowered ankle exoskeleton that reduced the metabolic cost of walking below normal gait using only a spring and a passive clutch tuned to ankle biomechanics, showing that correctly shaped morphology can substitute for actuation and computation alike [14].

A complementary strand treats embodiment as a physical precondition for learning, since intelligence acquired jointly through learning and morphological evolution outperforms learning over a fixed body, indicating that morphology and control co-determine what is even learnable and that the body participates directly in cognition [20]. Perspectives on intelligence in soft robotics make the partition explicit, distinguishing physically embodied intelligence (from structure and passive dynamics), materially embodied intelligence (from responsive materials), and algorithmic intelligence (from an explicit controller), and arguing that capable soft systems distribute competence across all three [21]. Disembodied artificial intelligence locates competence almost entirely in an algorithmic layer consuming sensor streams and emitting commands, treating morphology and physical interaction as fixed boundary conditions, whereas embodied intelligence treats structure, material, and interaction as design variables that shape what the system can do before any high-level processing occurs [22].

When a robot is worn, this picture acquires a feature absent from the literature that motivates it. The effective body is no longer the robot alone but the coupled human–robot ensemble, so that morphology, sensing, and control are shared across a biological and an engineered subsystem that are mechanically and informationally interpenetrated [23]. The human exceeds the environmental role assigned by the classical formulation and becomes an active, sensing, deciding, and adapting partner whose morphology and motor learning enter the system's intelligence budget; it is this asymmetry, absent from the embodied-intelligence accounts inherited from soft robotics, that the human-centric qualifier is introduced to capture.

## 2.2 HCEI for Wearable Soft Robotics

*We define HCEI as the capacity of a human–robot ensemble (a soft wearable robot and its wearer) to achieve assistance or augmentation by distributing intelligence across morphology, multimodal body-linked sensing, adaptive control, and the wearer's physiological and behavioral adaptation. Its objectives are defined in terms of human function, comfort, agency, and long-term well-being rather than the device's task performance in isolation.*

The definition is constructed from three commitments specialized to wearables: that autonomy and adaptivity arise from interaction among body, controller, and environment [22]. Second, the relevant body is the human–robot pairing, so morphology and perception are shared quantities [23]. Third, the human is treated as a co-equal agent (a design stance adopted here, not yet an empirically demonstrated property of current systems) whose goals govern a system whose remit extends beyond limb trajectories.

Imported definitions are insufficient precisely where wearables are most demanding, because the soft-robotics formulation is silent on the wearer as an adaptive agent, thereby omitting a partner that learns, fatigues, and exercises preference over days and weeks [21]; the artificial-intelligence formulation is silent on morphology and physical interaction, and on the privacy, agency, and life-context that become first-order constraints once a system is worn continuously [20]. HCEI aims to bridge these gaps. It retains the embodiment insight that competence is distributed across structure, material, and interaction, while adding the requirements (long-term human adaptation, agency, and contextual embedding) that a worn system cannot defer [23]. Assembling these commitments into a single criterion is a contribution of this review, and Section 3.3 develops its consequences for design sequence.

## 2.3 Fundamental Principles

The definition is operative only if decomposed into principles specific enough to serve as design and evaluation criteria; five are proposed. Morphological intelligence holds that a fraction of the assistive competence must be encoded in garment geometry, anchoring, and spatial compliance rather than delegated entirely to the controller. Anatomy-compatible load paths and passive elements then present the controller with a mechanically pre-conditioned problem [14], [18]. *Multimodal, body-linked perception* requires sensing that is continuous, multimodal, and physically proximate to the body (spanning kinematics, interaction forces, and physiology) embedded in the worn structure, as in inductive bimodal sensors reporting proprioception and tactile contact together or fluidic sensing in soft open-cell foams, with the distinguishing demand that perception include self-perception of the combined human–robot state together with exteroception [24], [25]. *Adaptive and anticipatory control* separates reactive parameter tuning from anticipatory behavior that predicts the wearer's intent and pre-shapes assistance before a movement unfolds, depending on intent and trajectory estimation of the kind now demonstrated through learned prediction of hand trajectories from egocentric observation [26].

The fourth principle, *human–robot co-adaptation and agency*, states that adaptation is bidirectional and unfolds over days and weeks. Human-centricity makes this mutual adaptation a requirement and adds an agency condition: the wearer must understand, shape, consent to, and override assistance, ensuring users retain ultimate control and never reducing the wearer to a passive payload [27]. The fifth, *contextual and societal embedding*, holds that a worn system is embedded in a life and a society, so that all-day comfort, sustainability across a long service life, privacy of the intimate physiological data the system necessarily collects, and compatibility with real workflows are constitutive requirements of HCEI, recasting the wearable as a personalized and sustainable artefact co-designed with the individual wearer. Recent syntheses already recognize personalization and sustainability as legitimate design objectives, and HCEI contributes by inserting them into a single dependency structure in which contextual and societal embedding both constrains and is constrained by the other four principles. Stated together, the five principles convert the definition into a checklist against which a candidate system can be assessed, and they recur as criteria in the grand-challenge and roadmap discussions that close the paper.

### 2.4 Relationship to Existing Paradigms

The construct's value is clearest when positioned against the paradigms it subsumes, beginning with classical assistive robotics, which, exemplified by conventional rigid exoskeletons, locates intelligence largely in the controller and treats the human as a payload moved along a reference trajectory; its design optimizes torque and joint kinematics, and morphology enters mainly as a frame to be aligned with the biological articulation [28]. Human-in-the-loop and adaptive control advanced this by tuning assistance to the individual and the moment, yet where they optimize a performance objective without co-designing morphology and human factors, they remain only partially embodied and largely silent on comfort, agency, and long-term adaptation [9]. Disembodied, AI-first approaches occupy the opposite extreme, concentrating competence in a model mapping sensor streams to commands while treating body and physical interaction as secondary, forgoing the competence available in structure and material [20]. Against these, HCEI is distinguished by where intelligence is located (distributed across morphology, sensing, control, and the wearer's adaptation), by the role it assigns the human (a partner whose agency and long-term well-being are design objectives), and by the horizon over which it is evaluated (comfort, trust, co-adaptation, and longitudinal outcome rather than immediate task performance). The paradigm is therefore subsumptive rather than oppositional, incorporating classical control, human-in-the-loop optimization, and learning-based models as components while requiring each to be co-designed with the morphology and human factors the standalone paradigms leave implicit, and judges the result by criteria they do not apply [22].

## 3. A Unified Framework for Soft Wearable Robotics: The PCAA Framework

HCEI supplies the lens through which this review reads the field; the PCAA framework provides its organizational structure. The preceding section argued what a soft wearable robot should become; this section explains how such a device is organized. PCAA decomposes a soft wearable robot into four coupled functions (Perception, Cognition, Actuation, and Augmentation) through which information and energy flow from the joined human–robot state to a measurable change in the wearer's function. The central claim is not that these four functions exist; every wearable system senses, decides, acts, and produces an effect. The claim is that they are *coupled*, that they must be reasoned about *together and from the outset*, and that a common design sequence in the field (which begins with an actuator and appends sensing, then control, then a use case after the fact) inverts the dependency structure of the problem it is trying to solve.

*We therefore define PCAA as the organizing framework of this review, within which any soft wearable robot is decomposed into four coupled functions (perception of the joined human–robot state, cognition that turns that state into a decided action, actuation that delivers it, and augmentation as the realized change in the wearer's function), which are mutually dependent and must be reasoned about together and committed concurrently from the outset.* PCAA is the only construct in this review designated a framework; it operationalizes the HCEI paradigm for analysis and design.

### 3.1 The Limits of Existing Classifications

Devices collected under "soft wearable robotics" are sorted by actuation principle, assisted joint, clinical application, or control strategy. Each scheme is internally coherent, yet none places morphology, function, intelligence, and human factors in a single frame. Reviews of active soft wearables typically partition the field by actuation mechanism, then cross it with target body segment and application domain, locating a device by what drives it and where it acts [9], [29]; systematic reviews of upper-limb devices add control- and readiness-based categories [30], [31]. Two limitations recur. First, morphology, function, and intelligence are treated as independent sorting keys, obscuring their interactions, so the literature can report a device's actuation type, target joint, and control strategy without expressing how its embodiment constrains the sensing and control it can support [32]. Second, human factors (comfort, donning and doffing, social acceptability, and the co-adaptation that unfolds over weeks) are discussed qualitatively but rarely admitted as an explicit axis, even though these same factors are widely recognized as the dominant barriers to translation. A taxonomy that merely adds a fifth bin cannot repair this; what is required is a frame in which the bins are explicitly coupled.

### 3.2 The Four Pillars as a Coupled Design Space

PCAA names the four functional pillars of an embodied wearable and asserts their mutual dependence. **Perception** comprises all sensing of the human and the environment (inertial measurement, load and force sensing, soft strain and pressure transduction, electromyography, and physiological monitoring), including the stretchable force sensing through which a soft exosuit observes its own interaction state [33]. **Cognition** comprises the layer that turns perceived state into action: finite-state and impedance control, intent estimation, and adaptive or learning-based policies, including the estimation of locomotion joint angles and moments that converts soft-sensor streams into personalized commands [34]. **Actuation** comprises the mechanical output (the actuation principle, force-delivery mechanism, placement, and control bandwidth), spanning the pneumatic, cable-driven, and shape-memory-alloy systems cataloged across the embodiment categories [29]. **Augmentation** comprises the realized effect on the human, whether restored or improved function, reduced effort, enhanced performance, or altered perception. This is the outcome by which the device is judged [35].

These pillars are not a pipeline that one walks once from end to end, since perception is meaningful only relative to the cognitive decisions it must inform. Cognition is bounded by what the actuator can deliver and what the perception layer can observe. The actuator's compliance and placement determine both the forces available for augmentation and the signals available for perception. In reverse, the augmentation target dictates what must be sensed and decided in the first place. A device's physical embodiment, ranging along a rigid-to-soft spectrum from garment-like textile exosuits with remotely sited actuation [13] through hybrid soft–rigid frames to hand-scale robotic gloves [36], [37], fixes the mass-distribution and force-routing boundary conditions on which all four pillars then depend. The four descriptive properties that prior reviews already measure (embodiment, functional objective, intelligence level, and human-integration profile) are retained as read-outs of where a device sits within the coupled PCAA space. They no longer serve as the primary organizing keys.

The four pillars also reorganize the field's recurring review themes onto a single spine, and it is along these pillars that the subsequent sections are arranged. Table 1 maps the themes onto the pillar each principally informs and to the section in which it is developed.

**Table 1. Mapping of the field's recurring review themes onto the four PCAA pillars.**

| Recurring theme in the literature | PCAA pillar it principally informs | Developed in |
|---|---|---|
| Multimodal on-body sensing, soft sensors, the Human Digital Signature | Perception | Section 4 |
| Intent estimation, learning-based and shared control, the wearable intelligence stack | Cognition | Section 5 |
| Soft materials, morphology, actuation mechanisms, and force delivery | Actuation | Section 6 |
| Restored or improved function, effort reduction, performance gain | Augmentation | Section 7 |
| Clinical validation, comfort, acceptance, and deployment at scale | Cross-pillar (human-integration read-out) | Section 8 |
| Coupled grand challenges and capability roadmap | Cross-pillar synthesis | Section 9 |

The intelligence dimension is itself ordinal and orthogonal to actuation principle, since cognition resides in the sensing and control wrapped around an actuator, not in the actuation type. We retain a five-level ladder (proposed here as an ordinal scale, with no claim to established-standard status) running from Level 0 (passive or structural competence residing only in the morphology of elastic garments and compliant supports), through Level 1 [38], Level 2 [33], and Level 3 [34], to Level 4 (human-centric embodied intelligence as a target state that no fielded soft wearable yet satisfies in combination). The ladder measures how much of the Cognition pillar a device actually instantiates, and it is the axis along which the field's frontier is presently advancing.

### 3.3 The Limitations of Actuation-First Design

The framework carries a normative consequence, because the pillars are coupled and the sequence in which they are committed during design determines which couplings can still be honored later. Actuation-first design (selecting a pneumatic, tendon, or shape-memory mechanism, then instrumenting it, then writing a controller, then identifying a population that might benefit) fixes the actuator's mass, bandwidth, and force-routing before the augmentation target has constrained them, so that the very boundary conditions on perception and cognition are set by a component chosen in ignorance of what it must serve. The recurring translation failures reported across the field share a common source in this inverted commitment order: assistance tuned on one cohort does not generalize across users and must be re-optimized in the loop for each individual [39], [40]. Comfort and donning burdens suppress adherence regardless of bench-measured benefit [41]. Intent detection is defeated when the physical interface perturbs the very signals it depends on. The corrective the framework enforces is to reason about all four pillars simultaneously at the conceptual stage. The augmentation objective and the human-integration constraints are stated first, perception and cognition are specified as the intelligence required to meet them, and actuation is selected last, as the downstream means of delivering a decided action. This is why the sections that follow place the intelligence core (perception and cognition) ahead of actuation, demoting the actuator from the origin of the design to its terminus.

### *3.4 Reading a Device's PCAA Engagement Profile*

The framework is operational, in that for any device one can state its embodiment category, functional objective, intelligence level, and human-integration profile, and trace which of the four pillars it materially engages, so the same system is describable both as a point in the descriptive space and as a perception-to-augmentation flow (Fig. 3). Figure 3 renders PCAA as a vertical stack with physical embodiment at the base and augmentation at the top. An ascending information path runs from perception through cognition, a descending energy path runs through actuation into the coupled body, and a closing feedback path returns augmentation to perception. Each exemplar carries a strip marking which pillars it instantiates.

Three mappings illustrate the read-out, of which the first is the biologically inspired lower-limb walking exosuit, which traces the full flow (inertial and load perception, gait-phase and force-tracking cognition, cable-mediated actuation, and a metabolic-cost augmentation) at an adaptive intelligence level with a moderate human-integration profile [13]. The soft robotic glove for grasp restoration is reactive-to-adaptive and high in human integration by virtue of its low mass and textile interface, running from grasp-intent perception through trigger or electromyographic cognition to pneumatic finger actuation, clinically validated after spinal cord injury [16], [36]. The shape-memory-alloy fabric-muscle suit is a suit-type robot in which embodiment and actuation coincide: its strip shows perception absent and cognition minimal (a direct thermal trigger), with actuation routed through the worn fabric and its human-integration profile set by the thermal and bandwidth limits of the embedded muscle [42]. Applied across the device population, the same procedure lets the sections that follow deepen one pillar at a time while keeping each device legible as a whole, and locates the field's open problems as specific gaps along the pillars.

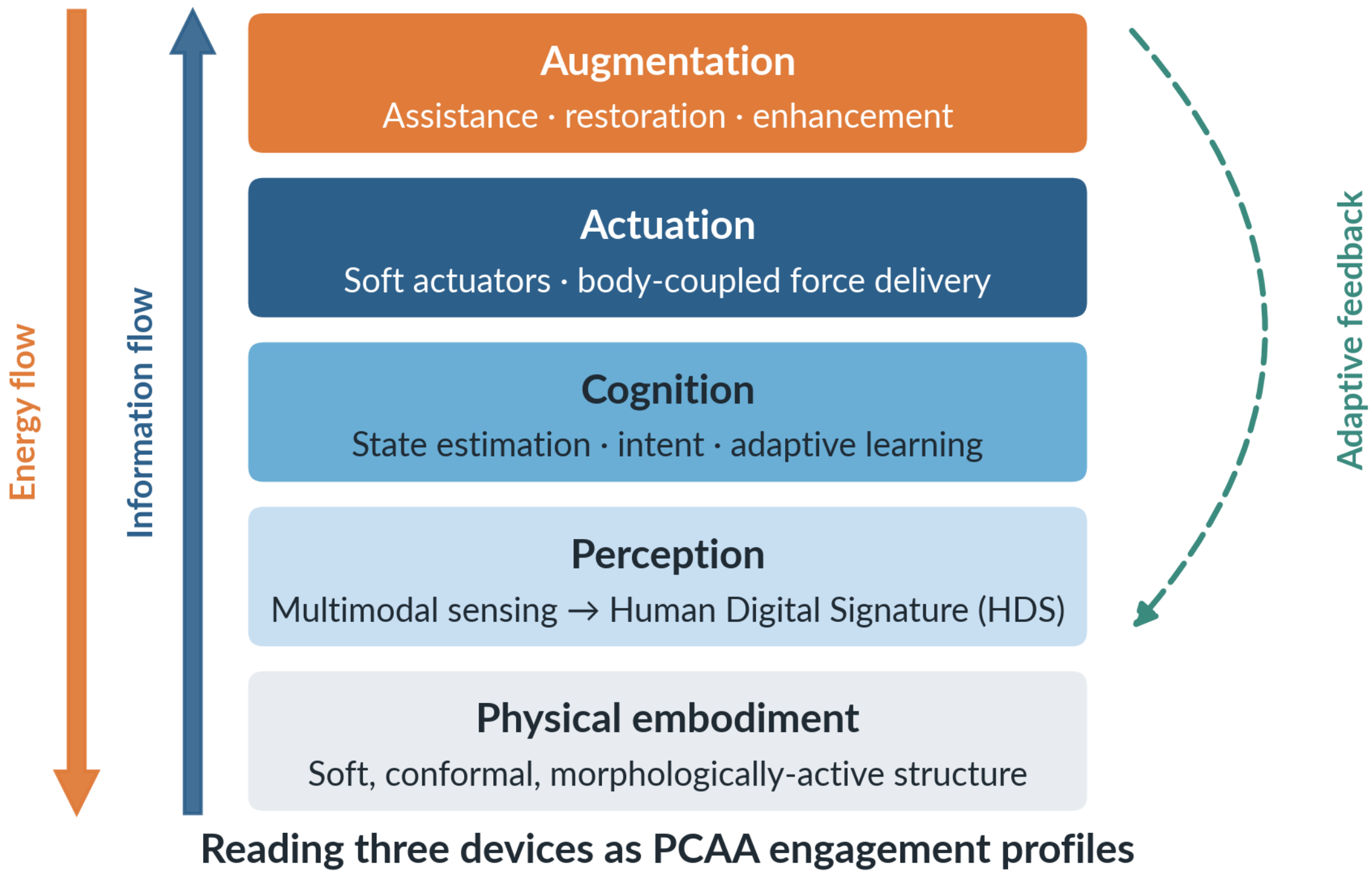

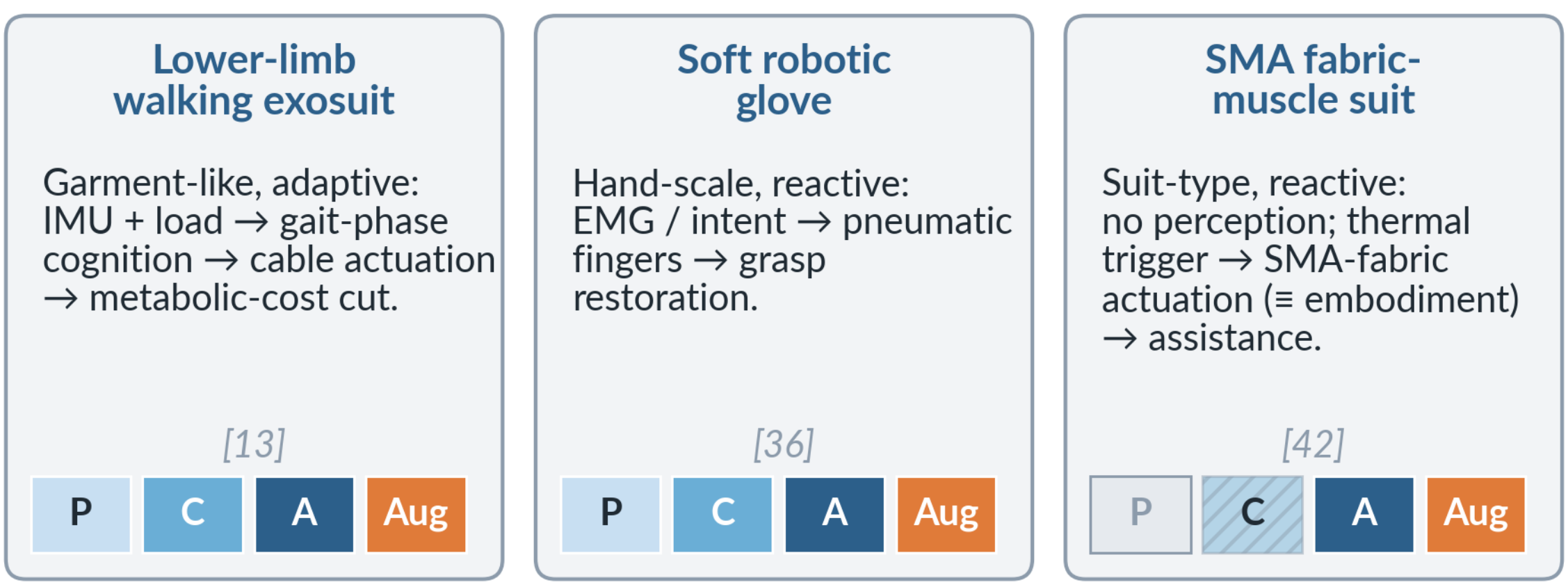

**Fig. 3. The PCAA framework as an integrative co-design lens.** The four functional pillars (Perception, Cognition, Actuation, Augmentation) are arranged as a vertical stack grounded in Physical Embodiment. Information flows upward from sensing to augmentation outcomes; energy flows downward from actuation into the coupled body; a closed adaptive feedback loop returns outcome data to the perception layer. Three representative devices are mapped to the framework to show how any soft wearable system can be located within it; the color-matched strip beneath each exemplar marks which PCAA pillars the device engages (gray = absent, hatched = minimal). The walking exosuit and soft glove trace the full perception-to-augmentation flow, whereas the shape-memory-alloy fabric-muscle suit has no dedicated perception and only direct-trigger cognition, with actuation coincident with embodiment.

# 4. Perception: Understanding the Human and Environment

Perception is the first pillar of the PCAA pipeline and the one on which the others depend for their inputs; a soft wearable can decide, act, and augment only as well as it can sense the coupled human–robot state it is embedded in. The wearer is an integral part of the closed-loop system: an active subsystem whose intent, effort, and fatigue must be read before any action is possible [43]. A wearable measures signals (joint angles, contact pressures, muscle potentials, and heart rate) but must act on latent states such as intent, effort, fatigue, comfort, and context. These states must be inferred through a many-to-one, context-dependent mapping: the same inertial trace changes meaning across terrain, and the same muscle potential changes meaning with fatigue [44]. Perception is therefore an inference problem layered on measurement, and this section follows the information as it ascends, from multimodal body-linked sensing of the human and the environment, through the fusion that turns isolated measurement into state estimation, into the Human Digital Signature (HDS) it composes, the personalized representation that the cognition pillar (Section 5) then consumes to decide and deliver assistance.

## *4.1 Sensing the Coupled Human–Robot State*

Much of a soft wearable's perception is built from the same compliant materials that form its body, since many of the stretchable polymers, functional fibers, optical waveguides, and conductive composites that make a wearable soft are simultaneously its sensing media, so structure and sensor are frequently the same element, the perception-side expression of the morphological-intelligence argument [45]. The signals fall into five modality classes. Kinematic and biomechanical sensing recovers how wearer and device move and load one another through inertial units, soft strain and capacitive sensors batch-fabricated into garments, optical waveguides, and tendon-anchor load cells, and is the modality on which most closed-loop soft wearables depend, directly shaping the assistance delivered [46], [47]. Bioelectrical sensing recovers neural and muscular activity, principally through surface electromyography from skin-compatible textile or dry electrodes, and high-density decomposition can resolve individual motor-unit activity in real time, pushing the channel from coarse activation envelopes toward the neural drive itself [48], [49]. Where electrode contact is the limiting factor, force myography offers a mechanical proxy for the same intent, since a band of force-sensitive resistors reading the expansion of forearm muscle groups has driven four-class hand-motion classification at approximately ninety-five percent real-time accuracy for control of a soft robotic glove [50].

Physiological sensing recovers internal state (cardiovascular activity, respiration, temperature, and biochemical markers) used primarily for personalization, comfort, and safety, as in clinically validated wearable ultrasound for continuous blood pressure and integrated arrays for in-situ perspiration analysis [51], [52]. Environmental sensing recovers the operating context (terrain, slope, obstacles, task setting) but remains underdeveloped, with terrain typically inferred from gait patterns because direct measurement remains uncommon, so the "environment" half of a wearable's perceptual problem is at present the weaker half [53].

Sensor fusion is where isolated measurement becomes state estimation and is the true bridge into cognition. Low-level fusion de-drifts and calibrates individual sensors, while high-level fusion produces the latent states the device must act on, as when combining electromyography with kinematics improves intent detection and soft strain with inertial data improves joint-angle and joint-moment estimation [33], [34].

### *4.2 The Human Digital Signature (HDS)*

The five signal modalities and their fusion supply the components of a personalized, multimodal, time-evolving representation of the wearer, the HDS (Fig. 4). The HDS is proposed as an abstraction over multimodal wearable data that encodes how a given person moves, activates, responds, fatigues, and interacts across sessions [43]. Its principal components follow from the modality classes: a baseline kinematic and biomechanical profile of habitual movement and interaction loading, a strength, activation, and fatigue profile from the bioelectrical and physiological channels, a physiological-response profile spanning cardiovascular, thermoregulatory, and recovery dynamics, and a behavioral and environmental-exposure profile of routines, tasks, and contexts [48], [52].

The Human Digital Signature provides several tangible benefits. By encoding how an individual moves, fatigues, and recovers, it enables controllers to personalize assistance through adaptive force, timing, and impedance, preserve these settings across sessions, and detect physiological changes before they progress to injury [34], [51].

Figure 4 renders the HDS as a layered model comprised of a base layer of multimodal body-linked sensing across the five classes, a middle layer of fusion and state estimation that converts raw signals into the four component profiles, and an upper layer comprising the personalized signature itself, with continuous refinement after calibration. The HDS is a section-level model that operationalizes the Perception pillar, not a second framework alongside PCAA. It makes explicit how perception connects to everything downstream. The HDS supplies the substrate for the human digital twin and scalable personalization at deployment (Section 8). It also supports adaptive cognition that anticipates intent and shapes assistance (Section 5), and it finally affects trust because a device adapted to this wearer is experienced differently from one that imposes a generic model. Its continuous, longitudinal maintenance for anticipatory intervention is a projection that remains to be established.

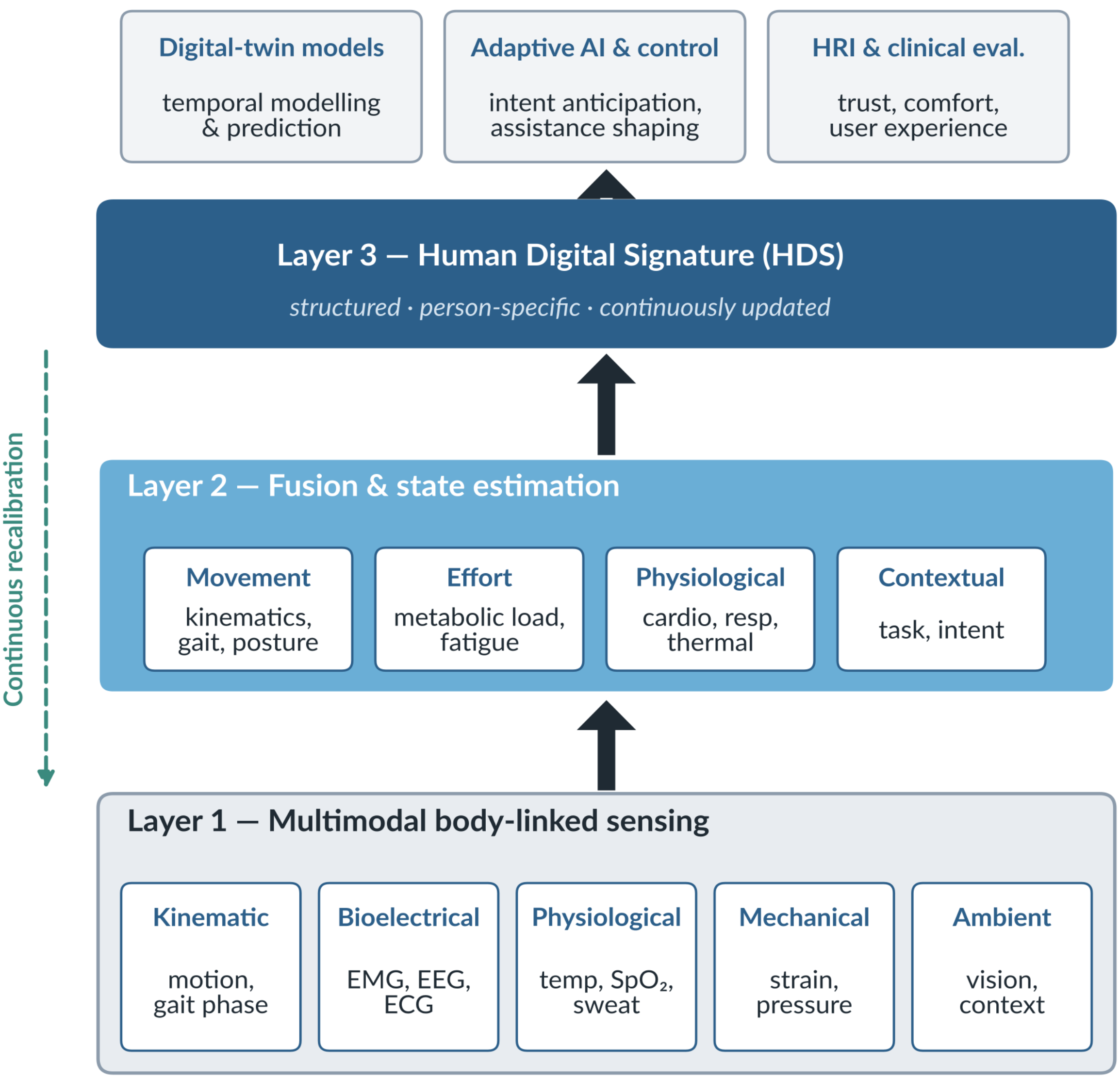

**Fig. 4. The Human Digital Signature (HDS): a model operationalizing the Perception pillar.** Five multimodal sensing classes (Layer 1) are fused into four component profiles via state estimation (Layer 2) to produce the HDS: a structured, person-specific representation updated continuously rather than fixed at calibration (Layer 3). The HDS operationalizes the perception side of PCAA and supplies the personalized substrate that the cognition, personalization, and deployment analyses consume.

### *4.3 The Challenge of Robust Perception*

The sensing described so far is most accurate in the controlled conditions under which it is characterized, and the defining perceptual problem of a fielded soft wearable is that daily use is not those conditions. Soft transducers exhibit hysteresis, creep, and drift; textile electrodes shift and lose contact as the garment moves; and motion artifacts corrupt the physiological channels precisely when the wearer is most active. And a state estimate that is accurate in one session degrades across a day of wear as the interface loosens and the skin condition changes [44]. Low-level fusion de-drifts and recalibrates individual sensors, but the deeper difficulty is distribution shift (the statistical relationship between signal and state moves as the wearer fatigues, sweats, and re-dons the device, so that perception cannot be validated once and assumed fixed). Robust, self-calibrating, all-day perception is therefore an open requirement of the pillar and remains unsolved, and it sets a ceiling on how much the cognition, actuation, and augmentation stages above it can reliably deliver [53]. For PCAA, the implication is that perception bounds the other pillars most tightly, because every downstream decision inherits the ceiling set by signal quality and state-estimate stability, a system co-designed for robust, self-calibrating, all-day sensing (with sensing co-designed before the actuator is fixed) is the precondition for cognition and augmentation that survive outside the laboratory.

# 5. Cognition: Intelligence, Learning, and Personalization

The second pillar, cognition, is the point at which a perceived state becomes a decided action. It consumes the HDS composed by perception and converts a state estimate into assistance, spanning structured, analyzable controllers and learning-based policies personalized to an individual. Because one person wears the device across many days, cognition must decide well in the moment while continuing to learn, personalize, and share authority over time [54]. This section follows that arc, starting from the learning that turns soft-sensor streams into estimates and intent, through the continual and personalized adaptation by which a controller tracks a changing wearer, into the hybrid architectures that make on-body learning safe, and out through the shared, co-adaptive control by which authority over movement is negotiated with the wearer.

## *5.1 Learning for Estimation, Prediction, and Intent*

Most deployed intelligence in soft wearables remains structured and task-specific (impedance, admittance, disturbance-observer, and robust feedback designs) with comparatively modest learning components for estimation and pattern recognition [54], [55]. Conventional machine learning serves two functions: regression maps soft-sensor streams to continuous variables such as joint moments and interaction forces, and discrete models classify activity and intent. Both are light enough for on-body use but prone to per-user overfitting and degradation under the distribution shift of all-day wear [34].
Deep models become attractive where the perception stage leaves dense, noisy, multimodal time series, learning temporal and cross-modal structure that hand-designed features miss. Recurrent and convolutional networks estimate joint moments in real time for exosuit control, multi-resolution networks estimate gait phase from inertial data, and subject-independent networks classify continuous locomotion mode for hip-exoskeleton control [34], [56], [57]. These methods are primarily used for intent prediction (estimating near-future states far enough ahead that actuation can be pre-shaped), which moves a system up the intelligence ladder from reactive assistance, timed after events occur, to predictive assistance timed to a forecast; robust prediction across the many transitions of daily life remains open [57], [58]. Such predictive, person-specific assistance is typically achieved through human-in-the-loop optimization, which searches a low-dimensional assistance-parameter space against a measured physiological objective and can reduce metabolic cost below fixed controllers within a session. Reinforcement learning is the principal present-day tool for closed-loop personalization, but subject to the constraints that exploration be safe on a body-coupled device and sample-efficient enough for the few minutes of data a wearer tolerates [39], [40],

[59], [60].

### 5.2 Continual Learning and Personalized Adaptation

Estimation and intent prediction are usually validated within a session, but the defining cognitive demand of a worn device is longitudinal. The same person changes from hour to hour and week to week, so a policy fitted once decays as fatigue accumulates, the interface loosens, and the wearer's own motor patterns adapt to the assistance, and personalization must therefore continue across sessions [34], [60]. This raises the problem of continual learning on the body (updating a per-user model as new data arrive without catastrophic forgetting of what already worked, and without unsafe exploration on a force-coupled limb), which is why present-day personalization relies on bounded, sample-efficient adaptation of a small parameter set; open-ended online retraining remains impractical [60], [61]. The human digital twin integrates these adaptations across time by turning the HDS from a static representation into a predictive engine. This review develops the twin in Section 8 as the mechanism through which personalization can scale at real-world deployment.

### 5.3 Hybrid Control and the Wearable Intelligence Stack

Because reinforcement learning and adaptive methods let a controller change its behavior online, they sharpen a tension a safety-critical wearable cannot avoid. Structured controllers achieve good performance with behavior that can be analyzed, bounded, and explained, as in the simplified non-smooth feedback control demonstrated for soft elbow assistance, whereas learning-based policies capture unmodeled dynamics and adapt more flexibly but are harder to verify, certify, and explain to a clinician or regulator [54], [62]. Hybrid architectures resolve this tension by ensuring that a model-based safety envelope bounds the region within which a learned component may personalize, gaining adaptivity without surrendering the predictability a body-coupled device demands [61]. Future directions include more speculative developments, which are stated here as projection. A wearable foundation model trained across many users, sensors, and tasks and personalizable by fine-tuning is becoming feasible on large wearable-sensor corpora, though none yet runs on a deployed soft wearable. Moreover, a user-facing personal agent holding the wearer's signature and mediating assistance, explanation, and data-sharing would operationalize the agency principle only if it never overrides the wearer [53], [63]. Fig. 5 renders cognition as the Wearable Intelligence Stack, a proposed layered view within the pillar. It comprises a material base encoding morphological intelligence, multimodal sensing that produces the HDS, fusion and estimation, cognition bounded by a model-based safety envelope, and an interaction layer for intent understanding and the prospective personal agent. The learning components sit inside the safety envelope to express the hybrid-architecture principle, and the foundation model and agent are marked as projected, so the stack maps where intelligence should be located in a human-centric system and does not describe a deployed system [61], [63].

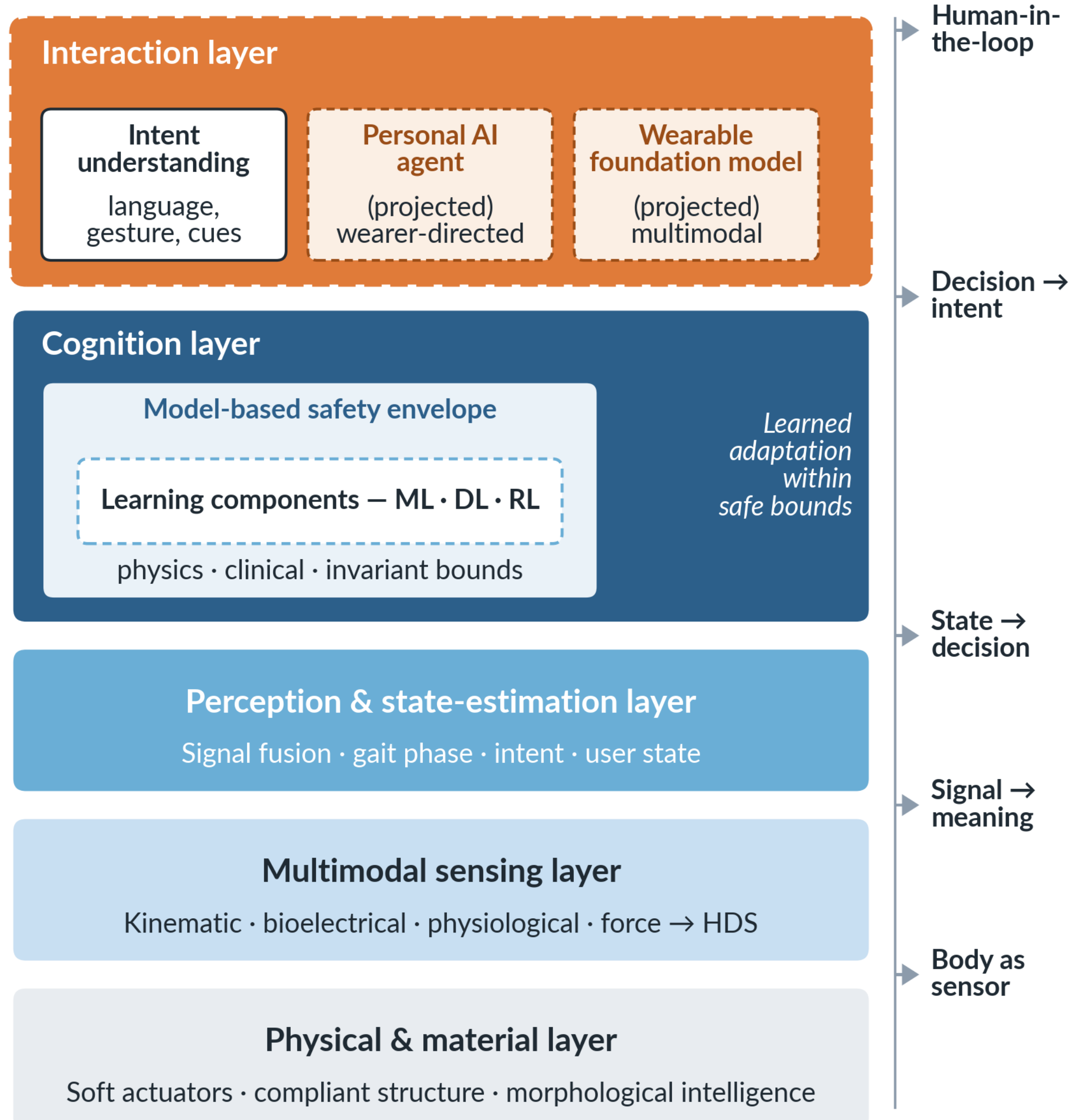

**Fig. 5. The Wearable Intelligence Stack: a layered view within the Cognition pillar.** Five layers organize the functional hierarchy of a human-centric soft wearable system, from the morphological intelligence encoded in soft materials at the base through sensing, perception, and cognition, to an interaction layer at the apex. Within the cognition layer, learning components (ML, DL, RL) are rendered inside the model-based safety envelope to express the hybrid-architecture principle that learned adaptation operates within fixed bounds. Upper-layer components (the personal AI agent and the wearable foundation model) are marked as projected to reflect their emerging status.

## *5.4 Shared Control, Co-adaptation, and Trust*

A wearable that senses, models, and adapts to its wearer is still only half of an assistive system. Its quality depends on participation in a shared loop of feedback, control, and adaptation through and on the body, making fit, timing, comfort, predictability, and trust first-order technical variables [53]. Feedback is the channel through which the device makes its state legible, and the wearable-haptics palette (vibrotactile, skin-stretch, force, thermal, electrotactile) is best organized by functional role: sensory substitution, guidance and error correction, immersive contact rendering, and confirmation that assistance is engaging as expected [64], [65]. Legibility lets the wearer participate actively in shared control, since authority over movement is distributed across user, controller, and device morphology, with systems using high-level intention detection to decide when to assist while a low-level controller executes the force, and soft wearables embodying shared control physically as well as algorithmically, since compliance and underactuation bound how strongly the device can dominate the limb [57], [66].

Once authority is genuinely shared, neither party stays static, because humans continuously optimize the energetic cost of walking and re-tune their gait in response to sustained assistance, so the operating conditions a controller was designed against shift as the user adapts; and a device must keep adapting across repeated calibration cycles [27], [60]. A direct consequence is that single-session evaluations are often insufficient, because some systems require a familiarization period before their benefit stabilizes, and the time course of co-adaptation is itself an outcome to be measured.

This loop stabilizes only if the wearer keeps adjusting to the device, and that willingness rests on trust, a technical property depending on mechanical transparency (the device does not feel obstructive when inactive), behavioral transparency (assistance is predictably timed), and informational transparency (the wearer can understand why assistance changed). Informational transparency is important precisely when the device adapts, since opaque adaptation can reduce acceptance even when it improves objective performance [65]. Acceptance is thus a core technical benchmark, not a soft outcome, because a device that is not worn cannot assist; comfort, donning burden, social acceptability, and real-world use time are translational variables of the same order as efficacy [41], [67]. When feedback, shared control, co-adaptation, trust, and acceptance hold together over time, wearer and device function less as operator and tool than as a single cooperative system. This is a condition of *shared intelligence*, proposed here as a synthesis construct in which authority, information, and adaptation are distributed across the human–device–context loop, and one of the intermediate stages between context-aware wearables and the fuller human–AI symbiosis treated later as a longer-horizon projection. In cognition, the framework's coupling is actively negotiated across wearer and controller, and because authority, adaptation, and trust are distributed between them, cognitive competence must be specified as a shared, bounded, and legible process (not appended to an actuator after the fact) if the pillar is to convert a state estimate into assistance that a person will accept over months.

# 6. Actuation: Delivering Intelligent Physical Assistance

Actuation is placed downstream of the intelligence core because the framework challenges the actuator-first approach. The energy-delivery stage of PCAA has its ceiling set by the material base beneath it and its realized value set by the perception and cognition stages above it, because intent detection, interaction-force estimation, and human-in-the-loop optimization each extract more from a given actuator than fixed schemes can [33]. In a soft wearable the actuator is not a modular force source that can be specified in isolation, as it can in a conventional manipulator. The material body fixes its operating pressure range, efficiency, backdrivability, and force density, and the actuator in turn fixes the device's form factor, tethering burden, skin interface, acoustic and thermal footprint, and the wearer's perception of comfort, so that the choice of actuation is inseparable from wearability itself.

## 6.1 Soft Materials and Morphological Intelligence

The material body sits at the base of the stack because it fixes the boundary conditions through which every higher function operates. In a soft wearable, the constituent materials simultaneously form a load path to the skeleton, a sensing medium, a comfort interface, and a participant in control. Three material families supply complementary primitives. Elastomers offer high deformability and body-conformal mechanics, dominating soft pneumatic chambers and stretchable sensing, with dielectric elastomers achieving electrically actuated strains exceeding one hundred percent [68]. Hydrogels offer tissue-like softness and biocompatibility for biointegrated interfaces, their historical brittleness addressed by toughened double-network designs [69]. Textiles and fibers are the most wearable-native platform, aligning with garments and distributed force transmission, as in the strain-programmable fiber-based artificial muscle whose contraction is set by its draw geometry [70], [71]. These families are best used in combination.

Above the constituent material, geometry becomes a programming tool, so that origami self-folding sheets and kirigami skins render a single sheet anisotropic, reconfigurable, conformal by fold or cut pattern alone. Architected materials generalize this to lattices, auxetics, and graded composites embedding tunable stiffness and even fluidic logic in structure, as in auxetic glove-type haptic interfaces and embedded fluidic sensing in soft open-cell foams [25], [72], [73], [74]. The same programming extends to the actuator itself, as in a hybrid plastic-fabric bending actuator whose trajectory is set by the arrangement of replaceable inflatable modules along a non-inflating spine, so that module count, module geometry, and the placement of flow blockages reshape the bending profile without any change to the controller [75]. This is where the materials axis meets the construct's first principle.

Morphological intelligence (the wearable analogue of morphological computation) delegates part of a system's behavior to its body, so passive compliance absorbs perturbation without control intervention, anisotropic textile load paths route force along anatomically compatible directions, and stretchable optical or strain media turn deformation into signal, each offloading a function the control stage would otherwise compute [18], [76]. A wearable with better morphology is frequently more intelligent in the embodied sense, because it reduces the estimation and control complexity the intelligence core must carry [23].

Emerging materials include self-healing elastomers, gait-powered textile energy harvesters, and epidermal electronics that conform to skin. Self-healing elastomers and gait energy harvesters remain benchtop demonstrations without evidence from all-day human wear, so wearable readiness is unproven [77], [78], [79].

## 6.2 Actuation Families and the Central Trade-off

Actuator selection is a constrained optimization over competing wearable objectives; there is no universally optimal actuator. The mechanisms delivering the largest, most controllable output tend to impose the heaviest tethering or least conformable interface. Meanwhile, those that integrate most naturally into a garment tend to be the most constrained in bandwidth, force, or efficiency [29]. Pneumatic actuation is the canonical soft family, coupling intrinsic compliance and large deformation with elastomeric and textile structures, and it dominates gloves and soft exosuits through segmented pneumatic-network bending, fabric-based actuators, and the McKibben artificial muscle; its barrier is that complexity shifts to the bulky, hard-to-mount pressure source, so truly portable pneumatics remain a projection [80], [81]. Within the same family, a rotary rather than a bending geometry carries pneumatic assistance to proximal joints, as an array of soft pneumatic rotary actuators driving a hip-flexion exoskeleton delivers 19.8 Nm at thirty degrees of flexion from 86 kPa and lowers measured muscle activation during a leg raise by 43.5 percent [82]. Hydraulic and electrohydraulic systems trade higher force density against fluid mass, with the HASEL class recasting hydraulic transfer as a fast, silent, electrically driven mechanism that nonetheless demands high voltages and remains at benchtop maturity [83], [84]. Vacuum-driven actuators collapse a structured void to contract or stiffen, intrinsically fail-safe and useful for grasp assistance and anchoring [85].

Tendon-driven actuation is a particularly consequential family in soft exosuits because it separates where force is generated from how it is transmitted, placing a heavy motor proximally while thin cables route tension along compliant garment paths, such as lower-limb walking exosuits and cable-driven shoulder exosuits. Its barrier lies in the transmission and interface, where friction, routing, and anchoring quality govern comfort and transmission [13], [86].

Shape-memory systems deliver compact, silent, garment-integrable actuation without pneumatic infrastructure, as in suit-type robots whose shape-memory-alloy fabric muscles make the garment itself the actuator, but their thermal principle yields slow response and poor efficiency [42]. Electroactive dielectric-elastomer actuators promise muscle-like deformation and silent operation but are constrained by high driving voltages and durability against sustained cycling, and magnetic actuation remains a niche localized-haptic option [1], [68]. The decisive cross-cutting lesson is that the interface and anchoring that route force into the anatomy are as decisive as the actuator generating it, since an excellent actuator delivering through a poorly anchored interface produces shear and transmission losses that undermine assistance [87].

## 6.3 Actuator Comparison Matrix

Table 2 is a comparison matrix that rates the actuation families against a ten-axis rubric using transparent ordinal levels; ordinal rating is a deliberate choice, since the underlying force-density, power-density, and bandwidth figures are reported under heterogeneous, non-standardized test conditions that do not support a defensible numeric comparison across families [29]. Primary force characterization supports that choice, since the usable output of a bending pneumatic actuator is bounded by yielding and buckling whose onset shifts with material and length, and since resolving that output at all required a revised measurement method, so that a single numeric figure of merit is not well posed across families [88].

**Table 2. Actuator comparison matrix: ordinal synthesis of soft-actuation families across ten wearable-relevant axes.**

| Family | Force/torque | Specific power | Bandwidth | Safety | Transparency | Portability | Textile integration | Noise / thermal | Durability | Best-fit domain |
|---|---|---|---|---|---|---|---|---|---|---|
| Pneumatic [80], [81] | Moderate–High | Moderate | Moderate | High | Moderate | Low (tethered source) | High | Low noise / low thermal | Moderate | Hand & upper-limb rehab, compliant assistance |
| Hydraulic [83], [84] | High | High | Moderate–High | Moderate–High | Moderate | Low–Moderate | Moderate | Low noise / moderate (HV) | Emerging | High-force; HASEL muscle-mimetic |
| Vacuum-driven [85] | Low–Moderate | Moderate | Low–Moderate | High (fail-safe collapse) | Moderate | Low (pump) | Moderate–High | Low | Moderate | Grasp assist, stiffening, anchoring |
| Tendon-driven [13], [86] | High | High (remote motor) | High | Moderate (interface-limited) | Moderate–High | Moderate–High | High (textile routing) | Moderate (motor) | High | Lower-limb gait, multi-joint upper-limb |
| Shape-memory [42] | Moderate | High (force-to-size) | Low | High | Low | High | High | Low noise / high thermal | Low–Moderate | Light assistance, posture, integrated garments |
| Electroactive (DEA) [68] | Low–Moderate | Moderate–High | High | High | High | Moderate | Moderate–High | Silent / low thermal | Low | Haptics, future high-integration muscles |
| Magnetic [1] | Low–Moderate | Moderate | Moderate | High | High | Low (field hardware) | Moderate | Low | Moderate | Localized haptic / biomedical |
| Hybrid & quasi-passive [38], [89], [90] | Tunable | High (clutched/ elastic) | Tunable | High | High (clutchable) | Moderate–High | High | Low–Moderate | High | Energy-efficient multi-DOF assistance |

*Legend.* Ratings are ordinal syntheses across representative wearable-scale demonstrations rather than quantitative rankings: **High** indicates repeated evidence of favorable performance under wearable-relevant constraints, **Moderate** indicates demonstrated but context-limited performance, and **Low** indicates a recurring limitation. Intermediate and hybrid ratings (e.g., Low–Moderate, Tunable) indicate application dependence; ratings are calibrated within each column to separate families and are not comparable as numbers across columns. "Emerging" marks a family whose wearable maturity is still early, and parenthetical notes flag the dominant caveat (e.g., HV = high driving voltage; tethered source = off-board power unit).

Three observations follow and recur. No family is dominant across the rubric, and the apparent leaders trade strengths against weaknesses on other axes, so actuator selection is intrinsically application-conditioned. One promising direction is hybridization and quasi-passive design, because clutches, series elasticity, and passive energy storage relax the portability-versus-output tension that constrains every pure family [38], [89], [90]. And the matrix's ordinal ratings should be read as a structured, application-conditioned comparison with no claim to settled numeric measurement. Here the design implication for PCAA follows directly from the section's placement, because no family dominates and each trades output against wearability on orthogonal axes, actuation is properly selected only after the augmentation objective and the intelligence core have specified the action it must deliver.

# 7. Augmentation: Expanding Human Capability

Augmentation is the apex of the PCAA framework. It is the integrative pillar at which perception, cognition, and actuation are judged not by whether function is restored but by whether capability is extended beyond an already-capable baseline. Restoring or assisting impaired function is only part of what soft wearable robotics can do; the same platforms serve human augmentation in work, sport, extreme environments, and hybrid human–robot systems, where the goal is to reduce fatigue, prevent injury, expand capability, or sustain performance in demanding contexts [91]. Three use cases can share hardware but differ in objective: rehabilitation restores or retrains lost function, assistance compensates for a persistent deficit or task burden, and augmentation improves capability, resilience, or efficiency beyond the unaided baseline in users who may not be clinically impaired. The boundaries overlap (aging workforces and return-to-work scenarios sit between assistance and augmentation), but the primary objective sets the design priorities, because augmentation shifts the figure of merit from clinical outcomes toward fatigue, throughput, ergonomics, endurance, and human–robot coordination, re-weighting every taxonomy axis toward endurance, ruggedness, and coordination rather than therapeutic dosing. This re-weighting makes augmentation domains a separate analytical problem even when the underlying technology is unchanged. It also changes how performance must be evaluated, since a versatile, portable exosuit that lowers the metabolic rate of walking and running in unimpaired users is augmenting [5], whereas the same hardware restoring a patient's lost capacity is rehabilitating, and the evidentiary standards and risk profiles differ accordingly.

## *7.1 Industrial and Logistics Work*

Industrial and logistics work is among the most developed non-medical application areas and carries the most direct biomechanical evidence. The scoping literature maps occupational exoskeleton use across healthcare workers, social care, and industry, documenting both the breadth of deployment and the unevenness of the evidence [92]. The core value proposition is reduced musculoskeletal load, lower fatigue, and improved occupational health, with maximal strength amplification outside the primary objective. The primary literature supports this for specific task classes: an industrial exoskeleton for overhead work shows biomechanical and metabolic effectiveness during sustained shoulder-elevated tasks [3]; a passive back-support exoskeleton reduces back load during lifting, with independent evaluation confirming physiological benefit in lifting and forward-leaning postures [4], [93]; and variable-stiffness systems reduce fatigue during squatting [94]. Across lifting, carrying, overhead work, and static postures the evidence favors localized load relief, yet workplace success depends on context-sensitive ergonomics and cannot be inferred from laboratory electromyography reduction alone. Thermophysiological burden can undermine all-day use even when assistance is biomechanically effective [95], and worker acceptance turns on perceived usefulness, comfort, and fit [67]. An emerging direction couples wearables with collaborative robots into bidirectional human–exoskeleton–cobot systems, where adaptation is shared across worker, worn device, and workspace robot, linking industrial augmentation directly to the shared-control argument of the intelligence core [96].

## *7.2 Construction, Field Work, Defense, Sport, and Extreme Environments*

Beyond the factory, design demands diverge by operating envelope, in that manufacturing benefits from repeatable workflows and devices tuned to recurring tasks, whereas construction introduces variable terrain and irregular tasks, increasing demands on portability, robustness, and environmental tolerance. In both settings, a wearable robot reduces strain while preserving human control of the job; the defensible claim is ergonomic load relief on specific tasks, not blanket productivity improvement [92], [96].

Agriculture and field work form an emerging domain in which field tasks combine repetitive bending, carrying, and long-duration labor but impose stricter demands than factories on weather tolerance, maintenance simplicity, and mobility over uneven terrain. Modularity, cleanability, and robustness dominate over control sophistication, and the domain is especially relevant to aging agricultural workforces and under-resourced labor contexts, but evidence specific to agriculture remains absent, making it a priority for future development [92].

Defense and high-load mobility are treated as design requirements anchored in the load-carriage problem; operational effectiveness remains unvalidated. Where a biologically inspired multi-joint soft exosuit reduced the energy cost of loaded walking and optimized hip–knee–ankle assistance reduced the metabolic cost of walking with worn loads, defense imposes stronger requirements for energy autonomy, ruggedness, and compatibility with protective equipment. Moreover, enhancement and militarization raise governance questions regarding ethics [97], [98].

In sport the emphasis falls on performance shaping, technique guidance, and injury prevention rather than strength enhancement, with freedom of movement and low perceptual interference being paramount. The soft-wearable sports literature is thin, so augmentation in sport is best framed around precision and adaptation, and is likely to benefit from the HDS and digital-twin constructs as a prospect that remains undemonstrated.

Space and extreme environments serve as a design stress case, imposing suit-related resistance, load-redistribution problems, and high penalties for fatigue or error, where low-profile, body-conformal assistance integrated with a pressure suit confronts the field's most challenging reliability, compactness, and thermal-control requirements at once [99].

## *7.3 Teleoperation and Human–Robot Systems*

A distinct branch of augmentation extends the wearer's reach by coupling the body to a remote or virtual agent so that capability is projected into environments the body cannot itself occupy. Soft, low-profile wearables are well suited to this teleoperation and telepresence mode because the always-on, unobtrusive capture it requires is exactly what compliant, body-conformal sensing provides. Flexible wearable sensors read the operator's motion and intent while the wearable-haptics palette developed for the cognition pillar renders remote or simulated contact back to the skin, closing a bilateral loop in which the human directs a distant manipulator and feels its interaction, an arrangement framed as using embodied intelligence to “reach the unreachable” [26]. The shared-control, legibility, and trust demands identified in the intelligence core are amplified in teleoperation because latency and imperfect remote perception widen the gap between intended and executed action. Teleoperation therefore remains an emerging augmentation frontier at an early, pre-validation stage and will draw on the same personalized, predictive cognition developed elsewhere in the framework [26], [91].

### *7.4 Cross-domain Synthesis*

Different augmentation domains optimize for different outcomes, with industry and logistics prioritizing fatigue reduction, injury prevention, and workflow compatibility, construction and field work emphasizing robustness and adaptability to variable terrain, and defense and extreme environments emphasizing endurance, ruggedness, and energy autonomy, anchored in load-carriage exosuit and exoskeleton energetics [3], [98]. Sport emphasizes responsiveness, precision, and minimal interference, whereas teleoperation emphasizes low-latency coupling and faithful bilateral feedback [26]. Set side by side, these differences show that augmentation reinforces the taxonomy axes (embodiment, actuation, intelligence, and human integration all shift with domain) and that non-medical augmentation broadens the field’s relevance while complicating evaluation, because success criteria differ sharply across applications and no single benchmark transfers cleanly between them, a complication forwarded to the benchmarking challenge taken up later [92]. The recurring lesson across all domains is that the evidence is consistently stronger for ergonomic load reduction than for productivity, and that three cross-cutting constraints (thermophysiological burden, energy autonomy, and user acceptance) recur in every domain regardless of objective, marking the augmentation pillar as the place where the framework’s human-integration demands are tested under the widest range of real conditions [67], [95]. The implication for PCAA is that augmentation supplies the objective from which the other three pillars must be designed backward. Because the figure of merit shifts with domain (fatigue and throughput at work, endurance and autonomy in the field, latency and fidelity in teleoperation), stating the augmentation target first is what allows perception, cognition, and actuation to be chosen to serve a defined, domain-specific capability.

## 8. Clinical Translation and Real-World Deployment

Technical progress has not translated cleanly into patient benefit, in that devices have moved past proof of concept yet clinical translation lags prototype innovation because evidence quality, standardization, portability, and implementation remain uneven. Stroke alone remains among the leading causes of long-term disability worldwide, so even incremental gains in gait or hand function carry substantial population benefit and a correspondingly high bar for evidence [2]. The governing distinction is between technical feasibility (whether the robot works) and clinical readiness (whether it produces meaningful outcomes in real patients, under realistic workflows, with acceptable burden, cost, and safety); a wearable robot must satisfy at least five conjunctive criteria (safety, efficacy, usability, portability, implementation feasibility), so strength on one cannot compensate for failure on another [30]. This section reads translation as the last and most challenging span of the PCAA pipeline, where a perceiving, cognizing, actuating, augmenting device must become a product a health system can deploy, and where the human digital twin serves as the mechanism by which personalization is made to scale, extending beyond its role in cognition.

## *8.1 Clinical Evidence by Application Area*

The most mature evidence is in gait applications, where the seminal demonstration that a soft exosuit improves walking after stroke established clinical plausibility, a subsequent multi-site trial moved the evidence toward multi-center testing, and randomized evidence is now accumulating, with a controlled pilot reporting gait gains in sub-acute stroke and a mechanistic trial showing that ankle-targeted resistance increases paretic propulsion [15], [100], [101], [102]. The comparison class matters, since rigid powered exoskeletons remain more suitable for severe impairment such as motor-complete spinal-cord injury, whereas soft exosuits suit milder deficits, fatigue reduction, and assistance outside constrained settings [12], [103].

Pediatric, older-adult, and neural-interface evidence is also emerging: a lower-extremity exoskeleton improving knee extension in crouch gait, a hip-assist robot improving walking efficiency in older adults, and a brain–computer-interface-controlled exoskeleton in pilot randomized testing [104], [105], [106].

Upper-limb evidence remains fragmented across shoulder, elbow, and hand, with a meta-analysis of portable upper-limb robots pooling support for benefit while underscoring protocol heterogeneity and small cohorts [107]. Hand-specific soft robots and gloves matter because hand impairment strongly constrains independence, and the primary record is substantial (soft gloves combining assistance with at-home rehabilitation, tendon-driven grasp designs, and fabric gloves assisting hand function after spinal-cord injury), yet only a single fully powered home-device pilot randomized trial stands out, a direct measure of how thin the high-quality base remains [36], [37], [108], [109]. That record now extends beyond spinal-cord injury to chronic stroke, where a bidirectional fabric-based glove improved object-manipulation scores across all eight participants in tasks simulating activities of daily living, although a cohort of that size holds the finding at case-series strength [110].

## *8.2 Translational Barriers*

Weak evidence design is a main bottleneck, since small single-center samples remain common, protocols differ widely in dose and comparator, outcome measures are inconsistent in ways that obstruct meta-analysis, and follow-up is often short, so future trials must report both clinical and wearable-specific endpoints such as don/doff burden, comfort, and real-world use time [30]. Wearable robots are safety-critical, exchanging force with vulnerable users; soft systems mitigate some hazards through compliance but introduce others (pressure concentration, anchoring slippage, material fatigue, and, as systems become adaptive, control safety), and regulatory maturity lags demonstration for systems whose behavior changes after deployment [111]. Even beneficial systems fail if setup time is long, therapist burden high, or the device does not fit routine care pathways, and reimbursement evidence remains far less developed than proof-of-concept studies [8]. Translation increasingly depends on what happens outside the laboratory, as shown by an independent four-week, exosuit-assisted, post-stroke community walking program in which adherence depends on comfort, trust, and interaction quality as much as assistive performance [41]. These barriers consolidate into a readiness scorecard (Table 3) that standardizes discussion across heterogeneous systems without ranking devices. It applies qualitative levels by application class because the literature does not support formal quantitative scoring. Grades combine cited trials with editorial synthesis, and the adaptivity column remains hedged because most fielded systems use static or coarsely adaptive personalization.

**Table 3. Clinical Translation Scorecard.**

| Dimension | Lower-limb gait (stroke, SCI, CP, aging) | Upper-limb (shoulder–elbow–forearm) | Hand / glove (soft) |
|---|---|---|---|
| Safety profile & adverse-event reporting | Moderate: force exchange well characterized; reporting inconsistent [12], [100] | Emerging: fewer standardized safety reports [112] | Emerging: compliance lowers severity; reporting sparse [108] |
| Evidence quality (sample size, design) | Moderate: multi-site and RCT pilots exist; large multicenter trials scarce [100], [101], [103] | Emerging: meta-analysis available but heterogeneous, small cohorts [107] | Low–Emerging: feasibility-dominated; one standout RCT [109] |
| Clinical efficacy on meaningful outcomes | Moderate: gait speed, propulsion, endurance gains shown [15], [102], [105] | Emerging: ADL/function gains, fragmented endpoints [113] | Emerging: grip/ROM/task gains, heterogeneous [36], [114] |
| Usability, comfort, don/doff burden | Emerging: portability improving; setup burden persists [41] | Emerging: comfort and donning remain barriers | Moderate: soft gloves favor comfort; anchoring/donning issues [37], [108] |
| Portability & home/community readiness | Emerging: community programs demonstrated, not routine [41] | Low–Emerging: mostly clinic-based [113] | Emerging: home pilots exist [109], [115] |
| Adaptivity & personalization maturity (hedged) | Emerging: adaptive and neural-interface gait control in research and pilot RCT [60], [106] | Low–Emerging: mostly fixed/EMG-triggered [113] | Low: calibration-level personalization [109] |
| Regulatory status / readiness | Emerging: some cleared rigid devices; soft systems behind [12] | Low: limited approvals [111] | Low: few market-ready systems [109] |
| Workflow integration & reimbursement fit | Low–Emerging: health-economic evidence thin [8] | Low: workflow fit underdeveloped [107] | Low: early; home models promising [115] |

Translational readiness by application class across eight dimensions, graded ordinally (Low / Emerging / Moderate / High). Grades synthesize the cited evidence with editorial judgement and are intended to standardize discussion, not to rank devices.

## 8.3 Manufacturing, Regulatory Readiness, and Commercialization

Even a device with proven efficacy and a workable personalization strategy reaches few users unless it can be manufactured, cleared, reimbursed, and supported at scale, and these deployment functions are markedly less mature than the device science. Manufacturing scalability is constrained by the very softness that gives these systems their interface advantage. Compliant, multi-material, garment-integrated constructions are harder to fabricate reproducibly and quality-assure than rigid assemblies, and sustainable production across a long service life remains an unsolved design goal. Regulatory readiness lags demonstration, and adaptive systems whose behavior changes after deployment strain a clearance model built for fixed-function devices, most visibly for the neurotechnology-coupled and pediatric systems still early in the evidence base [111]. Commercial viability then rests on reimbursement and health-economic evidence, far less developed than proof-of-concept studies, so even beneficial systems can stall for want of a payment pathway despite technical viability. Technology-readiness is uneven across the deployment stack (device efficacy outpaces manufacturing, regulation, and reimbursement), and closing that spread is as decisive for adoption as any advance in the pillars themselves.

## 8.4 The Human Digital Twin as a Route to Scalable Personalization

Personalization is easy to demonstrate for one wearer in one session but difficult to reproduce across a population in routine care. The human digital twin is proposed to bridge that gap: a dynamic computational model updated by wearable sensing that turns the HDS into a predictive engine for personalized assistance [116], [117]. Deployment at scale may benefit from a digital twin because personalization over weeks is a bookkeeping problem across time (joint mechanics, fatigue, recovery, task exposure, adherence, and the wearer's own adaptation each evolve on their own timescale and condition one another), and a twin holds these variables together in one continuously updated model. Its kinematic layer rests on physics-based musculoskeletal modeling that estimates internal joint and muscle states direct sensing cannot observe and, in real-time electromyography-driven form, runs during use, now fed directly by soft wearable sensing [34], [118], [119]. The personalized digital-biomarker approach then captures a person's expected baseline variation so that a deviation reads as a meaningful event against the person's own baseline, reframing variability from noise into structure [120].

Three fidelity levels (static personalization, the periodically updated adaptive twin, and the continuously coupled closed-loop twin) organize current practice, with most fielded soft-wearable systems between the first two, as shown for paretic-stroke gait and an early twin-coupled self-balancing exoskeleton [60], [117]. The defining difficulty is the tension between individual fidelity and population scalability under privacy, cost, and validation constraints, so continuous twin-coupled control and longitudinal predictive intervention remain emerging targets conditioned on validation and are not yet demonstrated functions [116].

## 8.5 Ethics and Governance

Soft wearable robots raise important ethical considerations. They sense intimate body data, influence movement, may adapt autonomously, and can shift agency, privacy, and responsibility in daily life, so ethical concerns connect directly to system architecture and rise with capability. The more personalized and predictive a wearable becomes, the more the governance of data, agency, and accountability matters [121], [122]. Privacy operates at three layers: raw-data privacy of continuous sensor signals, inferred privacy of model-derived states such as fatigue and impairment progression, and contextual privacy of the routines and locations revealed by long-term use, extending risk to bystanders and co-workers [123], [124]. The core risk lies in the creation of persistent, highly individualized body profiles (exactly what the HDS and the digital twin formalize), so the representation that makes personalization powerful also makes privacy a first-order constraint, and consent must cover retention, inference, model updating, and sharing, not merely collection [123]. Personalization also does not guarantee fairness; data, sensor, and deployment bias can degrade fit, intent detection, and assistance for underrepresented users, making inequity a safety and efficacy issue felt physically [122].

Because wearable robots physically shape action, agency is central, and dependence is a governance variable whose effects depend on context. Behavioral shaping reads as therapeutic retraining in rehabilitation but as productivity pressure in augmentation, depending on whose goals it serves [67], [121]. Ethics also links directly to the intelligence core through explainability, accountability, and safe delegation, so that black-box models undermine trust, and responsibility becomes difficult to assign across manufacturer, clinician, user, and adaptive model. Failures in a body-coupled system have direct physical consequences, so ethical-AI requirements overlap with safety engineering. Without explainability, shared intelligence can become opaque delegated authority in which the user retains nominal agency but cannot exercise it [122], [125]. Similarly, cybersecurity is inseparable from safety, because connected wearables and twin architectures enlarge the attack surface and security failures have informational and physical consequences, with privacy-preserving federated learning advanced as one architectural response that does not resolve the problem on its own [126], [127]. Governance should therefore follow, yet current regulation is fragmented across privacy, medical-device, cybersecurity, and emerging AI law. Adaptive twins complicate classification because a device cleared once may behave differently after updates, so future governance should be risk-tiered and lifecycle-based: covering sensing, training, deployment, updating, and retirement as distinct regulated stages, with these gaps forwarded to the grand-challenge synthesis [121], [123].

# 9. Grand Challenges, Future Roadmap, and Conclusions

## *9.1 The Coupled Grand Challenges*

The field's real bottlenecks are system-level coupling problems, in that comfort depends on actuation and interfaces; intelligence depends on sensing and data quality; translation depends on benchmarking and regulation; and sustainability depends on materials, power, and lifecycle design [128]. A wearable robot comprises tightly coupled subsystems whose overall performance is often constrained by the weakest interface. Consequently, improvements in one subsystem may have limited impact unless they are matched by corresponding advances in adjacent components. Understood in this way, the grand challenges are cross-sections through the PCAA stack that span multiple pillars. Wearability comes first, because many other advances fail if the system cannot be worn, and it is an emergent property (pressure distribution, thermal management, donning burden, social acceptability) that "soft" does not automatically solve [95], [129].
Energy autonomy is the defining bottleneck for the move from tethered laboratory systems to untethered platforms, resolving into power-source mass, actuator efficiency, and subsystem integration, with biomechanical and textile energy harvesting and entirely-soft autonomous robots offering early demonstrations, with no fielded solution yet [78], [130], [131]. Reliable intelligence is better defined as accuracy, robustness, safety, and explainability together than accuracy alone, because benchmark success is not real-world reliability across daily use [128]. Digital-twin fidelity is the bottleneck between promising personalization and scalable deployment, and its defining difficulty is the tension between individual fidelity and population scalability under privacy and cost constraints. Trust and shared-control reliability, manufacturing scalability, clinical adoption, and the twin problem of benchmarking and regulatory harmonization complete the set, each integrative and translational in scope [8], [30].

These challenges produce a future research agenda of fifty open questions in the Supplementary Material. Five clusters organize them: wearability and human integration; energy and untethered systems; sensing, intelligence, and reliable adaptation; digital twins and personalization; and translation, benchmarking, and governance. Five are especially consequential for the intelligence core. How can assistance be personalized with under five minutes of calibration while remaining robust across days? What standardized metric best captures mechanical transparency during prolonged, multi-task use and exposes limitations missed in a single instrumented session? How should adaptive controllers detect and fail safely when their state estimates degrade? How should digital-twin fidelity be validated in the specific context of wearable assistance? And how should adaptive, post-deployment-updating devices be regulated and benchmarked across their lifecycle [116], [128].

## *9.2 A Capability Roadmap*

The review's evidence resolves into a staged roadmap from today's assistive soft wearables toward human-centric embodied intelligence. It is a capability roadmap and does not predict dates, since classification follows functional capability and human integration independently of chronology, so two devices of the same age can occupy different generations. Five generations are proposed as the review's own capability periodization, stated as projection. Generation I, assistive wearables, is the current foundation, offering task-specific, mostly reactive assistance with rule- or phase-based control and personalization limited to fit and calibration, in which soft wearables proved their core value of compliant assistance with lower mechanical constraint than rigid exoskeletons [9]. Generation II, context-aware wearables, adds situational awareness and richer state estimation so assistance varies with task phase or detected activity [61]. Generation III, personalized wearable intelligence, is the first true threshold, the move from context-aware to person-aware systems that maintain individualized models and adapt across sessions, connecting to the HDS, with feasibility anchored in optimized human-in-the-loop assistance that reduces metabolic cost through individualized optimization [98], [132].

Generation IV, human digital twins, extends personalization into predictive, model-based coordination in which a dynamic representation of the wearer anticipates state change and individualizes therapy; most current systems show precursors, so it is an emerging target whose milestone is validated twin updates that improve prediction or intervention quality, not the mere presence of a model. Generation V, human–AI symbiosis, is an original framing presented as a long-horizon projection; it is a persistent cooperative system with shared, bidirectional, multi-timescale adaptation in which human agency is preserved through explainability, trust, and bounded autonomy [122]. Figure 6 renders the generations as ascending bands against coupled axes of functional capability and human integration. Dashed

Generation IV and dotted Generation V mark their predicted status, and a separate lane lists six cross-cutting enablers (materials and morphology, energy autonomy, sensing and intelligence, digital infrastructure, human integration, and translation pathways) to show that every stage requires simultaneous co-maturation.

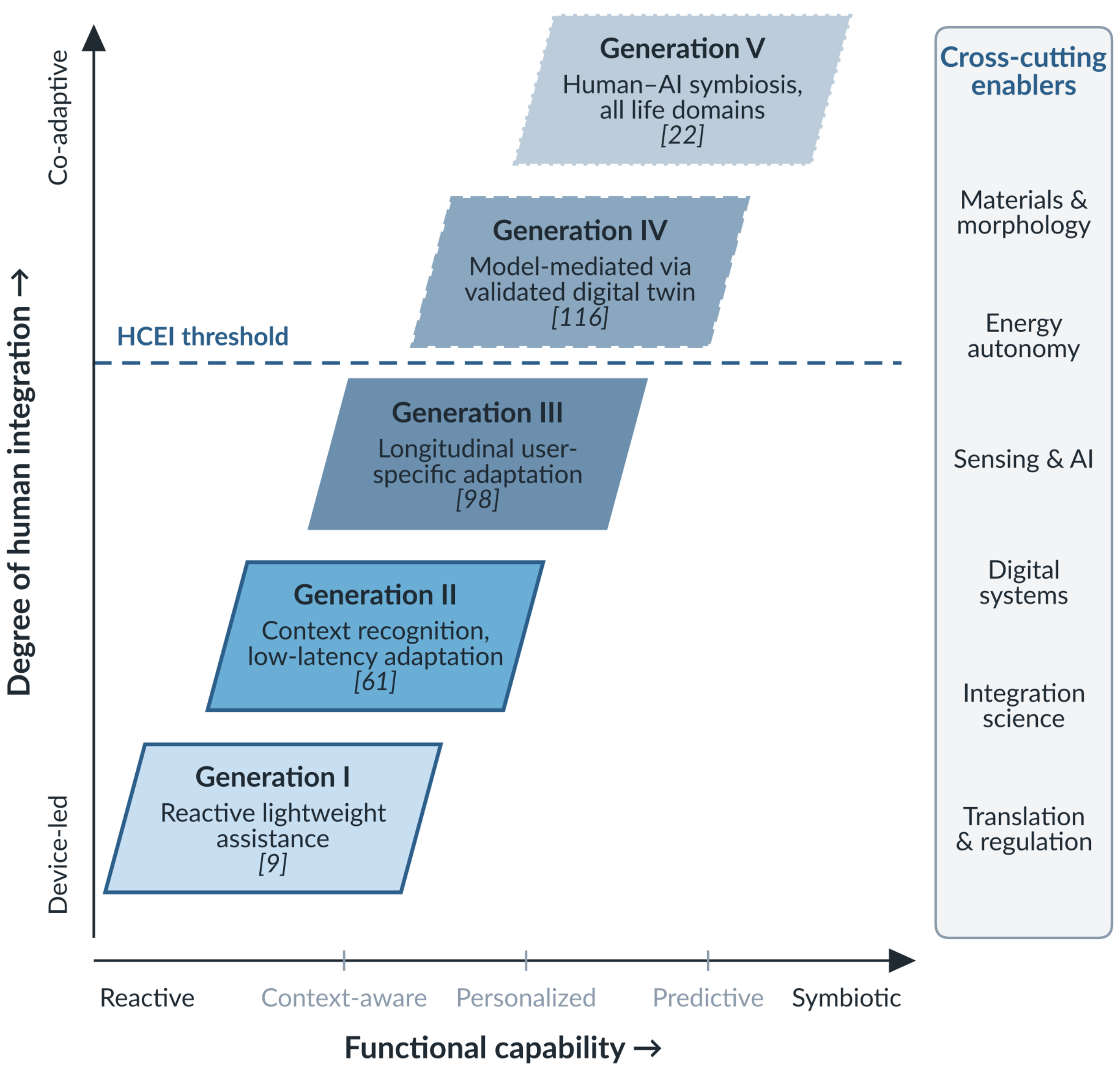

**Fig. 6. Roadmap to Human-Centric Embodied Intelligence across five capability generations.** The x-axis tracks functional capability from reactive assistance (Generation I) to human–AI symbiosis (Generation V); the y-axis tracks depth of human integration from device-led to co-adaptive. Generations IV–V are rendered with dashed and dotted borders respectively to indicate their prospective, hedged status. The dashed HCEI threshold at the Generation III/IV boundary marks the first point at which a wearable system achieves longitudinal, user-specific adaptation. Six cross-cutting enablers (right) must advance

simultaneously across all generations.

### 9.3 From Actuator-First to Human-Centric Design

The central design perspective proposed in this review can be summarized as a contrast between two design paradigms. Table 4 sets the actuator-first sequence the field has largely inherited against the human-centric order that HCEI and PCAA prescribe, across the dimensions the preceding sections developed; the contrast concerns when each decision should be made and does not diminish the importance of actuators.

**Table 4. From actuator-first design to human-centric embodied intelligence (HCEI) design.**

| Design dimension | Actuator-first design | HCEI design (this review) |
|---|---|---|
| Starting point | A chosen actuation mechanism | The augmentation objective and the individual wearer |
| Order of commitment | Actuator → sensing → control → use case | Objective and human constraints → perception and cognition → actuation |
| Role of the human | A payload moved along a reference trajectory | A co-adaptive partner whose agency and long-term well-being are design objectives |
| Locus of intelligence | Concentrated in the controller | Distributed across morphology, sensing, cognition, and the wearer's own adaptation |
| Where morphology enters | A frame to be aligned after the mechanism is fixed | A programmable part of the intelligence budget, specified up front |
| Figure of merit | Bench-measured torque or task performance | Comfort, trust, co-adaptation, and longitudinal outcome |
| Evaluation horizon | A single instrumented session | Days to months of real-world wear |
| Characteristic failure mode | Assistance that does not generalize; an interface that defeats its own sensing; abandonment despite bench benefit | Open problems located as gaps along specific pillars and resolved by co-design |

### 9.4 Limitations of This Review

Several limitations bound the claims made here, and stating them plainly is part of reading the review correctly. First, the methodology is integrative and narrative. Literature was selected for conceptual relevance through a purposive process, so the account synthesizes the field's trajectory without attempting an exhaustive device catalog or quantitative meta-analysis. Second, purposive selection introduces the possibility of literature-selection bias; coverage is weighted toward primary papers and review literature that carries the conceptual thread, so relevant studies (particularly negative results and gray literature) may be under-represented. Third, soft wearable robotics and its artificial-intelligence methods are advancing faster than a review cycle. Frontier claims about foundation models, digital twins, and adaptive on-body learning are therefore snapshots that will date quickly. The review responds by emphasizing durable organizing principles and presenting its forward-looking constructs as projections that require validation. Fourth, the synthesis depends on the available published evidence, whose base is itself uneven (small cohorts, heterogeneous protocols, inconsistent outcome measures, and few longitudinal deployments), so the readiness judgements assembled here inherit that uncertainty and form a structured orientation whose judgements remain provisional. These limitations qualify the review's conclusions without undercutting its central argument, which concerns how the field should be organized; quantitative device ranking lies outside its remit.

### *9.5 Conclusion*

Soft wearable robotics has crossed the threshold at which its most challenging problems are no longer mechanical but integrative. The field can build compliant actuators, conformal sensors, and adaptive controllers, but turning them into a system that a particular person will wear, trust, and benefit from over months remains the binding constraint. This review has argued that the most productive way to organize that integration is to read the field through human-centric embodied intelligence and the PCAA framework, treating perception, cognition, actuation, and augmentation as a single coupled design problem addressed concurrently from the augmentation objective and human-integration constraints. Placing the intelligence core ahead of actuation is a corrective design choice, and the recurring failures of the field (assistance that does not generalize, intent detection defeated by its own interface, and bench-validated devices abandoned in daily life) are its downstream symptoms. The supporting constructs developed here are offered as proposed organizing devices, labelled as such, that make the field's progress legible and its open problems locatable along specific pillars. The roadmap toward genuinely personalized, predictive, and cooperative wearables states the conditions each stage must satisfy. It does not forecast inevitable arrival, and every transition remains gated by a coupled grand challenge and a cross-cutting enabler. The field's next decade may depend less on stronger actuators than on whether soft wearable robots can be made wearable, trustworthy, efficient, scalable, and governable in real life; it is the human, not the device, around whom that progress is best measured.

## Acknowledgments

This research received no specific grant from any funding agency in the public, commercial, or not-for-profit sectors.

## AI-Use Disclosure

The authors used large language models to assist with drafting, restructuring, and copyediting of the manuscript text: Opus 4.7/4.8 (Anthropic), Sonnet 4.6/5 (Anthropic) and GPT 5.6

Sol/Terra (OpenAI). After using these tools, the authors reviewed and edited the content as needed and take full responsibility for the content of the publication. The tool was not used to generate, analyze, or interpret data, to produce original scientific claims, or to select or fabricate references.

## Conflict of Interest

The authors declare no conflict of interest.

# Supplementary Material

## *Table S1. Top 50 Open Research Questions in Soft Wearable Robotics*

### Cluster A: Wearability and Human Integration [129]

1. What standardized metric best captures “mechanical transparency” during prolonged, multi-task daily use rather than in a single instrumented session?
2. How can pressure at soft-tissue anchoring points be distributed to prevent injury without sacrificing force transmission?
3. What thermal-management strategies allow all-day wear of fluidic and electroactive wearables under occupational heat load?
4. How should donning and doffing burden be quantified and bounded for unsupervised home use?
5. What objective markers predict long-term device abandonment from short-term wear data?
6. How does social acceptability of visible wearables vary across age, culture, and setting, and how can design reduce stigma?
7. What is the minimal interface footprint that preserves assistance while restoring natural freedom of movement?
8. How can comfort be co-optimized with actuation efficiency rather than traded against it?
9. What materials provide skin biocompatibility over months of continuous contact and perspiration?
10. How should wearability be specified as a system-level requirement that constrains actuation, sensing, and control jointly?

### Cluster B: Energy and Untethered Systems [78], [130]

11. What power-source-to-assistance ratio defines a practically untethered multi-joint wearable?
12. How can actuator efficiency be improved enough to make all-day untethered fluidic assistance feasible?
13. To what extent can biomechanical and textile energy harvesting offset on-board power demand during real activity?
14. How should assistance strategies be designed to minimize energy per unit of delivered benefit?
15. What control architectures reduce sensing and computation energy without degrading adaptation?
16. How can on-board control and power be embodied softly without reintroducing rigid components?
17. What duty-cycle models best predict real-world battery life across heterogeneous daily use?
18. How can self-powering subsystems be integrated without compromising comfort or washability?
19. What is the achievable mass budget for an untethered exosuit that assists multiple

joints?
20. How should energy autonomy be benchmarked across devices with different assistance goals?

## Cluster C: Sensing, AI, and Reliable Adaptation

21. How can intent detection remain robust to sensor-placement change and signal drift across days?
22. What validation protocols establish reliability across users, tasks, and contexts rather than within one cohort?
23. How should adaptive controllers detect and fail safely when their state estimates degrade?
24. What level of explainability is required for users versus clinicians versus regulators of adaptive wearables?
25. How can multimodal fusion be made resilient to the loss or corruption of individual sensing channels?
26. What metrics distinguish genuine longitudinal adaptation from short-term performance fluctuation?
27. How can learning-based estimation generalize across body types and movement styles with limited per-user data?
28. What is the safe envelope of control authority for an artificial-intelligence-driven wearable in unstructured daily life?
29. How can model updates after deployment be validated without re-running full clinical evaluation?
30. What shared benchmark tasks would make intelligent-wearable results comparable across laboratories?

## Cluster D: Digital Twins and Personalization [132]

31. How should digital-twin fidelity be validated for wearable assistance rather than for clinical diagnosis?
32. How can assistance be personalized with under five minutes of calibration while remaining robust across days?
33. What minimal sensing set supports a useful adaptive twin without intrusive data collection?
34. How can twin submodels operating at different timescales be kept interoperable and synchronized?
35. What update-reliability guarantees are needed before a twin may influence control or therapy decisions?
36. How can high-fidelity personalization be delivered without amplifying inequity across resourced and under-resourced users?
37. What are appropriate quality metrics for a closed-loop wearable digital twin?
38. How can twins generalize a person's model across changing tasks, contexts, and recovery states?
39. What governance keeps a person's twin private while still permitting model improvement?
40. How can predictive intervention be evaluated for benefit and safety before routine deployment?

## Cluster E: Translation, Benchmarking, and Governance [123], [128]

41. What shared outcome set would harmonize clinical and wearable-specific endpoints across trials?
42. How should comfort, don/doff burden, and real-world use time be standardized as reportable trial endpoints?
43. What multicenter study designs are feasible given the heterogeneity of wearable systems?
44. How should adaptive, post-deployment-updating devices be regulated across their lifecycle?
45. What health-economic evidence would justify reimbursement for soft wearable assistance?
46. How can benchmarking account for task, user, interaction mode, and adaptation strategy simultaneously?
47. What implementation-science models best fit wearable robots into therapist and home workflows?
48. How should governance handle the dual-use tension between workplace safety monitoring and surveillance?
49. What risk-tiered framework appropriately classifies body-coupled, data-rich, adaptive wearables?
50. What certification would evaluate longitudinal use quality, not only instantaneous actuator output?

## *Table S2. Representative Device-by-Device Timeline of Wearable Robotics*

| Era | Representative system | Contribution |
|---|---|---|
| I: Mechanical assistance (≈1960–2000) | MIT-Manus [10] | End-effector robot-aided neurorehabilitation |
| | BLEEX [11] | Load-carrying lower-extremity exoskeleton |
| | HAL [133] | Cybernics-based whole-body assistive exoskeleton |
| II: Soft robotics revolution (≈2000–2015) | ReWalk [12] | Powered exoskeleton restoring ambulation after motor-complete SCI |
| | Bio-inspired soft exosuit [13] | Textile lower-limb walking assistance |
| | Passive ankle exoskeleton [14] | Metabolic-cost reduction with only a spring and clutch |
| III: Intelligent wearable systems (≈2015–2025) | Ankle soft exosuit [15] | Improved post-stroke walking; move to clinical testing |
| | Portable exosuit [5] | Reduced metabolic rate of walking and running |
| | Fabric-based soft glove [81] | Fully fabric bidirectional hand assistance/rehabilitation |
| | Soft glove, grasp restoration [16] | Soft glove grasp restoration after spinal cord injury |
| IV: Human-centric embodied intelligence (≈2025 onwards, prospective) | Personalized, sustainable wearables [134] | Personalization and sustainability as design objectives |